\documentclass{article}

\usepackage{arxiv}
\usepackage{array}

\usepackage[utf8]{inputenc} 
\usepackage[T1]{fontenc}    
\usepackage{hyperref}       
\usepackage{url}            
\usepackage{booktabs}       
\usepackage{amsfonts}       
\usepackage{nicefrac}       
\usepackage{microtype}      
\usepackage{lipsum}
\usepackage{graphicx}
\usepackage{array}
\title{Sentiment Analysis on Indian Indigenous Languages: A Review on Multilingual opinion mining }

\author{
  Sonali Rajesh Shah \\
  School of Computing \\
  Dublin Business School\\
  Dublin, Ireland \\
  \texttt{10402114@mydbs.ie} \\
   \And
 Abhishek Kaushik \\
School of Computing \\
  Dublin Business School\\
  Dublin, Ireland \\
  \texttt{Abhishek.kaushik@dbs.ie} \\
  \\
}

\begin{document}
\maketitle

\begin{abstract}
An increase in the use of smartphones has laid to the use of the internet and social media platforms. The
most commonly used social media platforms are Twitter, Facebook, WhatsApp and Instagram. People
are sharing their personal experiences, reviews, feedbacks on the web. The information which is
available on the web is unstructured and enormous. Hence, there is a huge scope of research on
understanding the sentiment of the data available on the web. Sentiment Analysis (SA) can be carried out
on the reviews, feedbacks, discussions available on the web. There has been extensive research carried
out on SA in the English language, but data on the web also contains different other languages which
should be analyzed. This paper aims to analyze, review and discuss the approaches, algorithms,
challenges faced by the researchers while carrying out the SA on Indigenous languages.
\end{abstract}

\keywords{Indian; Sentiment Analysis; Indigenous Languages; Machine Learning; Deep learning; Data; Opinion Mining; Languages.}

\section{Introduction}

SA is the process of extracting the opinions of people and use it to understand the people’s attitude, reactions expressed on the web regarding the various issues in the world and is also known as opinion mining. Nowadays with the increasing use of the internet a lot of information is available on the web which is about the different products, movies, books, technologies etc. People express their views, opinions etc on the different products,services,books etc on the web. For e.g. customer has bought a smart phone, as soon as the customer starts using the phone, he/she gives the feedback about whether they liked the phone, which features they liked or disliked. This type of reviews or feedback from the customers or users have become a boon to the industry. These views can help the industry or a company to improve their services i.e. if the reviews are negative then the aspects can be improved and if the reviews are positive, then that aspect can be kept in mind while creating a newer version of the service.

According to the authors Medagoda et al. \cite{medagoda2013comparative} there has being a continuous research going on in the English language but the research carried out in the indigenous languages is less. Also, the researches in indigenous languages follow the techniques used for the English language but this has one disadvantage which is, techniques have properties which are specific to a language. Hence It is really important to understand and analyze Indigenous language data because it can give meaningful insights to the companies. For example, India and China have world's largest population and are rich in diverse languages, analysing these indigenous language will be useful to companies because they have large share of users in India and China. In the current study, the types of languages i.e. indigenous languages and code mix languages are discussed prior to the approaches, methodologies used by the researchers and challenges faced by them.

\subsection{Indigenous Languages}
Indigenous languages are the languages that are native to a region or spoken by a group of people in a particular state. It is not necessarily a national language. For e.g. Irish, Tibetan, Spanish, Hindi, Marathi, Gujarati, Telugu, Tamil are the indigenous languages.
\subsection{Code Mix Languages}
Code-mixing is mixing two or more languages while communicating in person or over the web. Code-mixing is basically observed in the multilingual speakers. Code-mixed languages are a challenge to the sentiment analysis problem. A classic example of the code-mix language  is Hinglish which is combination of English and Hindi words present in a sentence. Hinglish is widely used language in India to communicate over the web. For e.g. movie review in Hinglish is “yeh movie kitni best hai.. Awesome.” In this sentence movie, best and awesome are English words but the remaining words are Hindi words, so the language identification becomes the first step in code mix languages followed by the SA which indirectly increases the overhead for the researchers and becomes time consuming process.

The remaining paper is structured as follows. Section II explains about the the process carried out in SA. Section III describes about SA levels and the different work done in each level. Section IV is about the current trending techniques in Natural Language Processing(NLP). Section V describes about the data-sets used by the researchers. Section VI explains about the SA techniques and the work done by the researchers using the different techniques. Section VII is about the challenges and limitations faced by the researches. Section VIII is the discussions and analysis about the papers been studied. Section IX is conclusion and future scope. 

\section{Sentiment Analysis Process}
\begin{figure}[h]
\includegraphics[width=\textwidth]{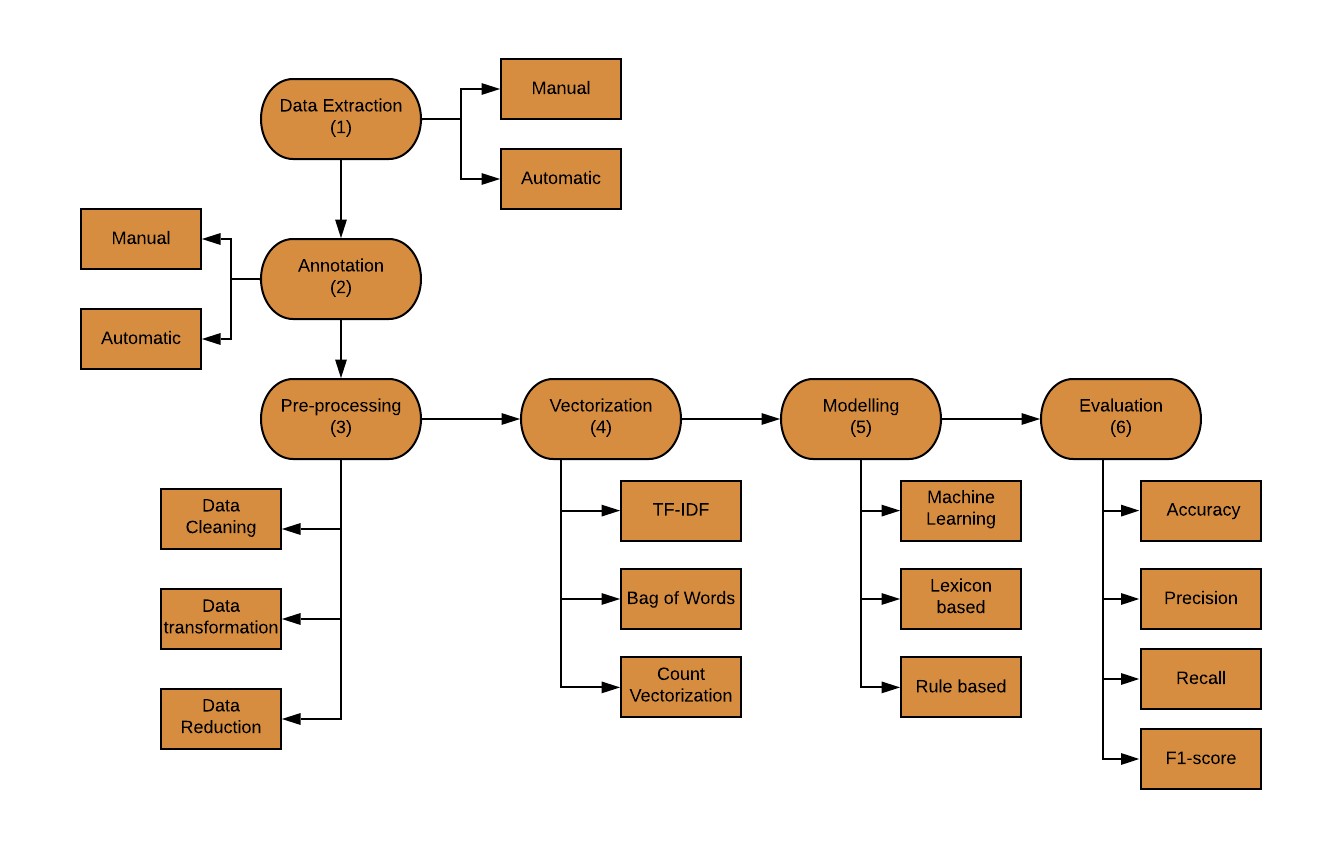}
\caption{Sentiment Analysis Process.}
\label{SA Process}
\end{figure}

The process of SA is carried out in 6 major steps which are data extraction, annotation, pre-processing, feature extraction, modelling, evaluation. Figure \ref{SA Process} shows the steps in the SA task and the explanation of each step is as follows.

\subsection{Data Extraction}

The first step of any SA task is data extraction. The data can be extracted either manually or automatically. Different web scraping algorithms help in automatically extracting the data from the web. One of the popular web scraping technique is Text pattern matching, which extracts only the information which matches the search criteria mentioned in the algorithm. Also, different Application Programming Interface (API) offered by social media platforms like Twitter, YouTube, Facebook etc. help in the data extraction process.
\subsection{Annotation}

Once the data extraction step is completed it is important to label the data. Annotation is process to add comments, observations, notes, questions related to the data at a particular point in the document. Labeling is a part of annotation and is used to classify the data as positive, negative or neutral. Labelling can be carried out manually or automatically. Majority of the researchers have done manual labelling on the dataset \cite{kaushik2016anatomy, kaushik2016comprehensive}.
Data collected from the web is raw and unstructured. It is essential for the researchers to carry out the pre-processing step which is as follows.

\subsection{Pre-processing}
Pre-processing is the process of converting the raw and unstructured data into the understandable and structured form. There are 3 major steps which are involved in  pre-processing which are data cleaning, data transformation, data reduction. Each step is explained as follows.

\subsubsection{Data Cleaning}
In this step the missing values and the noisy data is handled. Missing values in the data can be handled by filling the missing values manually or by finding the attribute mean or probability values. Noisy data can be due to data collection , data entry errors.Noisy data can be handled by using clustering algorithm. In clustering algorithm similar data is grouped together to form one cluster and the noisy data which is usually an outlier lies outside the clusters.

\subsubsection{Data Transformation}
Data is sometimes not in the suitable form for mining process therefore some type of transformation is required. Normalization, attribute derivation are ways of data transformation. Normalization is a process of scaling the data values in a specific scale ( 0 to 1 , -1 to 1). Attribute derivation is process of extracting the data from multiple attributes and creating new attribute. For e.g. age can be a derived attribute from the date of birth of customer.

\subsubsection{Data Reduction}
\begin{figure}[h]
\centering
\includegraphics{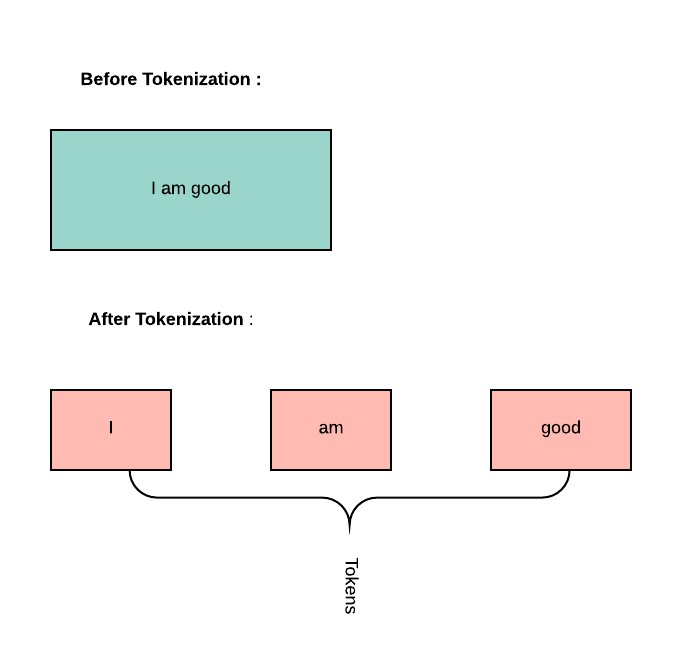}
\caption{Tokenization}
\label{tokenization}
\end{figure}

Data which is available on the web is huge. In order to process the data lot of efforts and time is required. There are some attributes in the data which are not that important and can be removed. Data reduction process can be carried out using attribute selection, numerosity reduction technique. Attribute selection is a process of selecting only the important and relevant attributes from the dataset and discarding the remaining attributes. Numerosity reduction stores the model of the data instead of the whole data.
There are different pre-processing techniques used by the researchers and the most common ones are tokenization, stop-word removal, Parts Of Speech Tagging (POS), stemming and lemmatization. Tokenization splits the data into individual words known as tokens \cite{kaushik2015study}. Tokenization can be explained using Figure \ref{tokenization} which is as follows.

Stop words are frequent words used in the sentences. Removal of stop words will not effect the sentiment polarity. Common stop words for English language are “is”, “was”, “there”, “that”, “they”,” he”,” she” etc. POS tagging is the technique where the words are tagged based on the parts of speech they are present in . For e.g. “ She is beautiful” for this sentence the POS Tagger will tag words as follows ‘She’- pronoun , ‘is’- verb , ‘beautiful’- adjective \cite{kaushik2015study}. 

Stemming is the process to reduced words to the root form by removing the suffix and the prefix present in the word. For e.g. “occurring” is the word the stem form of it is “occur” because the suffix “ing” is removed from it. One disadvantage of stemming is that sometimes the words do not have any dictionary meaning.

Lemmitization solves the problem of stemming. It first tries to find the root form of the word and then only the prefix and suffix of the words are removed. For e.g “leaves” is the word. The stem form of it is “leav” and the lemmitized form of it is “leaf”.

Different feature extraction techniques can be applied on this pre-processed data which is explained in detail as follow.

\subsection{Data Vectorization}

\begin{figure}[h]
\centering
\includegraphics{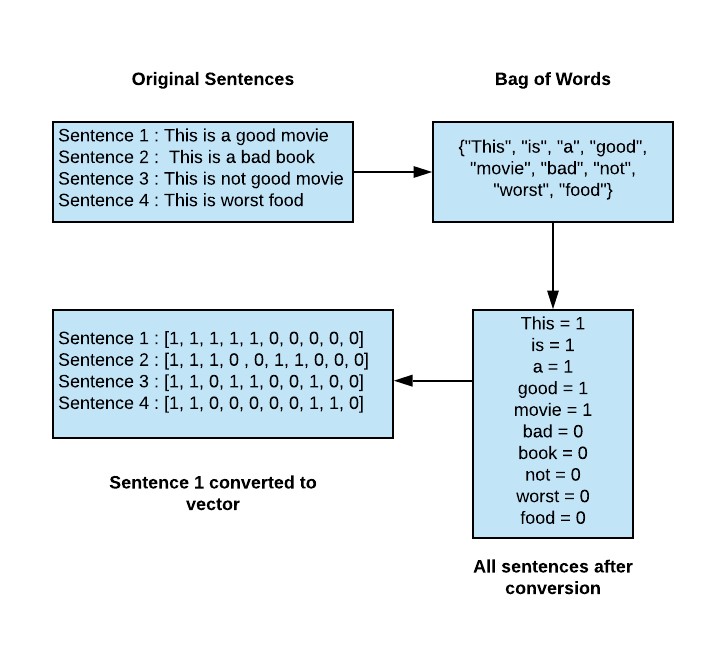}
\caption{Bag of Words Steps}
\label{BOW steps}
\end{figure}

Text vectorization is the process of converting the textual attributes into the numeric format. Machine learning algorithms usually work with the numeric data and hence there is a need to convert the textual data into the numeric or vector format. The most common vectorization techniques are bag of words, Term Frequency and Inverse Term frequency (TF-IDF) and count vectorizer. Bag-of-Words (BOW) is the most common vectorization technique. In this technique the pre-defined list of words i.e BOW is maintained and the words in the BOW are compared to the sentences. If the word in the sentence is present in the BOW list, it is marked as 1 else it is marked as 0. The vector created is of the size of the BOW. Figure \ref{BOW steps} explains the BOW in detail. 

TF-IDF is the very common feature extraction technique. It is a statistical measure to find how important the word is in the document. Term Frequency (TF) calculates the occurance of the word in the single document by the total number of words in the document, where as inverse term frequency (IDF) tries to find how important the word is in all documents \cite{schutze2008introduction}. 

Statistically TF and IDF are represented in equations \ref{eq:TF} and \ref{eq:IDF} respectively.

\begin{equation}
TF = \frac{\:Number\:of\:time\:word\:appears\:in\:the\:document}{\:Total\: words\: in\: the\: document}\,.
\label{eq:TF}
\end{equation}

\begin{equation}
IDF = \frac{log_e(\:Total \:number \:of \:documents)}{ \:Number \:of \:documents \:which \:contains \:the \:word}
\label{eq:IDF}
\end{equation}

\begin{figure}[h]
\centering
\includegraphics{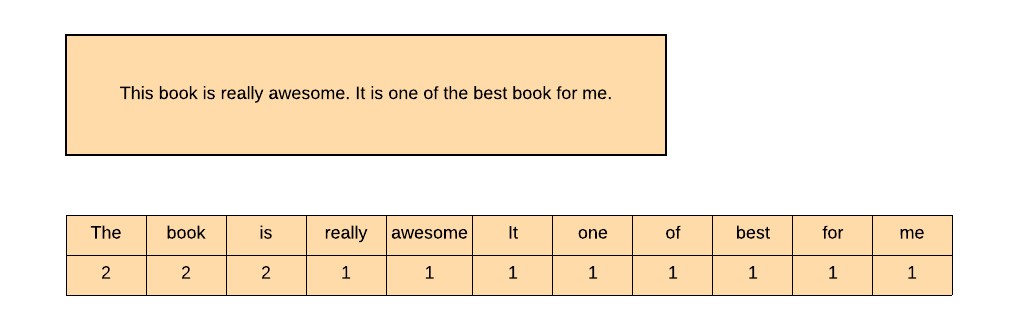}
\caption{Count Vectorizer}
\label{Count}
\end{figure}

Count Vectorization is a vectorization technique in which a document matrix is maintained. The document matrix contains the words present in each document with the frequency of occurrence of that word in the document. Figure \ref{Count} explains the count vectorization with an example. 

\subsection{Classification Techniques}

Classification of the data can be done by 3 approaches which are machine learning approach, lexicon based approach, rule based approaches. 

\subsubsection{Machine Learning approaches}

These are the approaches in which different supervised, unsupervised and semi-supervised learning algorithms are applied on the dataset to carry out the analysis and predictions. 

\subsubsection{Lexicon based approach}

In this approach the dictionary or corpora is used to carry out the SA task. In this approach the dictionary or the corpus words have polarity values assigned to each one of them. The words in the dataset are searched in the lexicon and if the word match is found the polarity of the word is assigned. For e.g the task is to find out the list of computer programming languages in the sentences which can be done using lexicon based approach by maintaining the predefined list of the programming language as a dictionary and then searching the words from the sentences in it.

\subsubsection{Rule based approach}

It is the traditional approach in which the set of rules are defined to carry out the SA task. For e.g the task is to find out the list of computer programming languages in the sentences. The rule developers scan the sentences and try to define rules which can perfectly predict the languages. Rule defined by developers is to extract all the capital words in the sentence except the first capital word. Test sentence is “Language above is Python”. The rule based approach will correctly identify the language but it will be failed when the sentence is “Java is programming language”. 

\begin{figure}[h]
\includegraphics{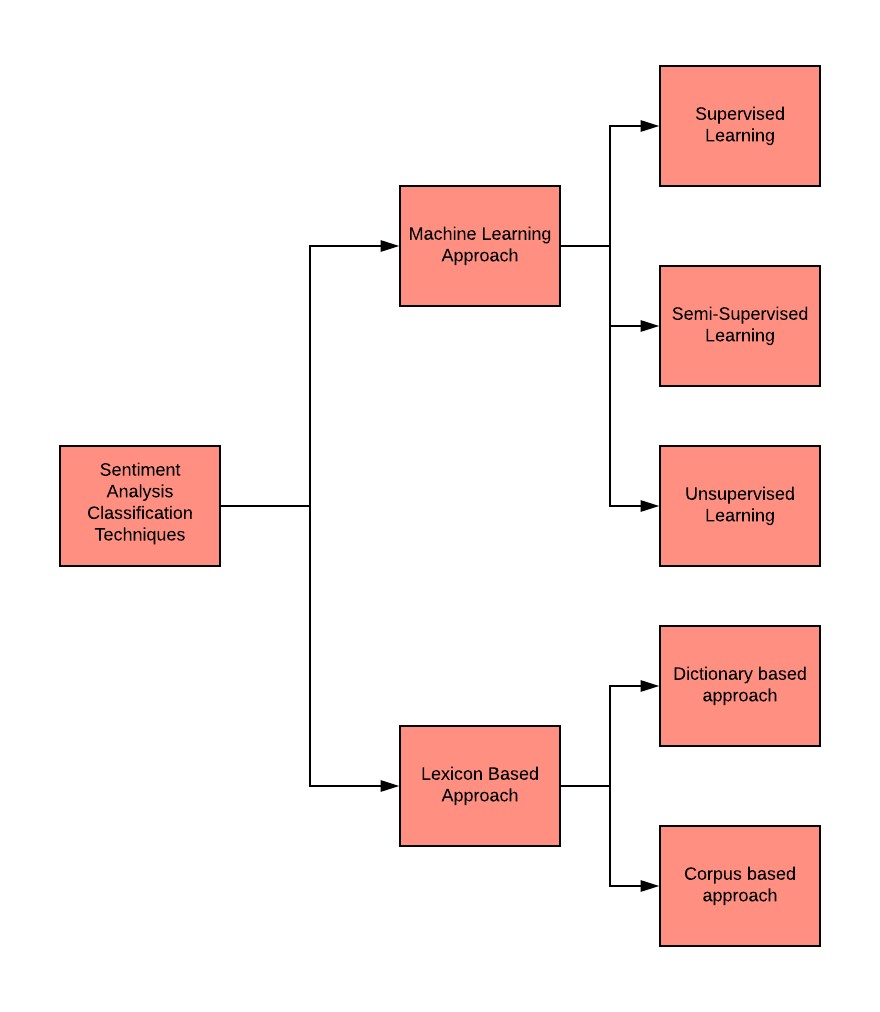}
\caption{Classification techniques}
\label{classification techniques}
\end{figure}

Figure \ref{classification techniques} represents the different classification techniques.

\subsection{Evaluation}
Once the model is validated and the results are available the different models are evaluated using different performance metrics. The most common performance evaluation metrics are accuracy , precision , recall , F1-score. 
\begin{itemize}

\item Accuracy:\\
It is the number of correct predictions over the total number of the instances of data \cite{schutze2008introduction}. 
\item Precision:\\
It is the number of the correct positive results over the total number of positive predicted results \cite{schutze2008introduction}. 
\item Recall: \\
It is number of correct predicted results over the total number of actual positive results \cite{schutze2008introduction}.
\item F1 score: \\
It is the weighed average of precision and recall \cite{schutze2008introduction}. 

Statistically accuracy, precision, recall and F1-score are represented in equations \ref{accuracy}, \ref{precision}, \ref{recall}, \ref{fscore} respectively.

\begin{equation}
Accuracy = \frac{\:TP\:+\:TN}{\:TP\: +\: TN\: +\: FP\: +\: FN}\,.
\label{accuracy}
\end{equation}

\begin{equation}
Precision  = \frac{\:TP}{\:TP\: +\: FP}\,.
\label{precision}
\end{equation}

\begin{equation}
Recall = \frac{\:TP}{\:TP\: +\: FN}\,.
\label{recall}
\end{equation}

\begin{equation}
F1-score = \frac{2\:*\:(Recall \: * \: Precision)}{(Recall \: + \: Precision)}\,.
\label{fscore}
\end{equation}

where , TP = Truly predicted positives, TN = Truly predicted negatives , FP = Falsely predicted positives , FN = Falsely predicted negatives.

\end{itemize}

\section{Sentiment Analysis Levels}

SA can be carried out at 3 levels. document level, sentence level and aspect level. 

\subsection{Document Level}

In this process the SA is carried out on the document or paragraph as a whole. Whenever a document is about a single subject it is best to carry out document level SA. Examples of document level SA datasets are speeches of the word leaders, movie review, mobile review etc.
   
SentiWordNet(SWN) is a opinion based lexicon derived from the WordNets. WordNets are the lexical database which consist of words with short definition and example. SWN consist of dictionary words and the numeric positive and negative sentiment score of each word. WordNets and SWNs are researchers common choice when carrying out SA on document level. Pundlik et al. \cite{pundlik2016multiclass} were working on multi-domain Hindi language dataset. The architecture implemented in the paper \cite{pundlik2016multiclass} contained two steps. Domain classification which was the first step was performed using ontology based approach. Sentiment classification being the second step was performed using HNSW and Language Model (LM) Classifier. There was a comparative study done on the results by the HNSW and HNSW + LM Classifiers. The combination of HNSW and LM Classifier gave better classification results as compared to HNSW \cite{pundlik2016multiclass}.
    
The work by Yadav et al. \cite{yadav2019semi} showed that SA for the mix-Hindi language can be performed using three approaches. The first approach was to perform classification based on neural network on the predefined words. Second approach used IIT Bombay HNSW. Third approach performed classification using neural network on the predefined Hindi sentences. The approaches in \cite{yadav2019semi} are explained in detail as follows. The first approach maintained the positive and negative word list. The mix-Hindi words were converted into pure Hindi words and were searched in the positive and negative list which was created manually. If the word was found in the positive word list the positive word count was incremented and if the negative word was found the negative word counter was incremented. In second approach instead of the positive and negative word list the HNSW was used remaining all the steps were same as in the first approach. In third approach seven features were created and applied on the sentences. The features are as follows, to find the frequency of the word, adjective, noun, verb, adverb, total positive polarity and negative polarity of the sentence. These features were send to the neural network for testing and the polarity of the word was detected. After the comparison of all approaches it was found that the second approach had the best accuracy which was 71.5\%.
   
Ansari et al. \cite{ansari2018sentiment} introduced an architecture for two code mix languages Hindi and Marathi. The architecture included language identification, feature generation and sentiment classification as major steps. Hindi and English WordNet’s and SWNs were used as there was no SWN for Marathi. The Marathi words were first translated into English and the sentiment score of the English words were found and assigned to the words. Also, classification algorithms like Random Forest, Naïve Bayes, Support Vector Machine (SVM) were used for finding the polarity in the final step. Slang identification and emoticons were also crucial steps in the study. Slang are a group of words which are used informally and in a particular language. Emoticons are the representation of different facial expressions. SVM performed the best among all the algorithms with accuracy of 90\% and 70\% for Marathi and Hindi language. 

In the paper, Jha et al. \cite{jha2015homs} explains that there is a lot of research done in the English language for SA, but little for the Hindi language. The system developed by the authors carried out the SA in Hindi language using two approaches. In first approach, supervised machine learning algorithm Naïve Bayes was used for document classification and in the second approach, the parts of speech (POS) tagging was done using TnT POS Tagger and using the rule-based approach the classification of opinionated words was completed. 200 positive and 200 negative movie review documents are web scraping for testing the system. Accuracy of 80\% was achieved by the system.

\subsection{Sentence Level}

Sentence level SA identifies the opinions on the sentence and classify the sentence as positive, negative or neutral. There are two types of sentences, subjective and objective sentences which are required to be identified while performing sentence level SA. Subjective sentences carry opinions, expressions and emotions in them. Objective sentences are the factual information. Sentence level SA can be carried out only on the subjective sentences hence it is important to first filter out objective sentences. 

SWN is a most common lexicon-based approach used by the researchers. Haithem et al. \cite{afli2017sentiment} developed the Irish SWN whose accuracy was 6\% greater than the accuracy obtained by transliteration of the Irish Tweets into English language. The lexicon was manually created. The accuracy difference between the systems was because of the translation carried out into the English language \cite{afli2017sentiment}. Naidu et al. \cite{naidu2017sentiment} carried out the SA on Telugu e-newspapers. Their system was divided in two steps. First step was subjectivity classification. Second step was sentiment classification. In the first step the sentences were divided as subjective and objective sentences. In the second step only, the subjective sentences were further classified as positive, negative and neutral. Both the steps were performed using the SWN which gave the accuracy of 74\% and 81\% \cite{naidu2017sentiment}. 

Nanda et al. \cite{nanda2018sentiment} used the SWN to automatically annotate the movie review dataset. Machine learning algorithms Random Forest and SVM were used to carry out the sentiment classification. Random Forest performed better than SVM giving the accuracy of 91\%. Performance metrics used to evaluate the algorithms were accuracy, precision, recall, F1-score \cite{nanda2018sentiment}.

Pandey et al. \cite{pandey2015framework} defined a framework to carry out the SA task on the Hindi movie reviews. \cite{pandey2015framework} observed that the lower accuracy was obtained by using SWN as a classification technique and hence suggested using synset replacement algorithm along with the SWN. Synset replacement algorithms groups the synonymous words having same concepts together. It helped in increasing the accuracy of the system because if the word was not present in the Hindi SWN then it found the closest word and assigned the score of that word \cite{pandey2015framework}. In the study, Bhargava et al. \cite{bhargava2016sentiment} completed the SA task on the FIRE 2015 dataset. The dataset consisted of code-mixed sentences in English along with 4 Indian languages (Hindi, Bengali, Tamil, Telugu). The architecture consisted of 2 main steps Language Identification and Sentiment Classification. Punctuations, hashtags were identified and handled by the CMU Ark tagger. Machine learning techniques like logistic regression and SVM were used for language identification. SWN’s of each language were used for sentiment classification. The results of the implemented system were compared with the previous language translation technique and 8\% better precision was observed \cite{bhargava2016sentiment}. 

Kaur, Mangat and Krail \cite{kaur2017dictionary} carried out their SA task on Hinglish language, which is code mix language highly popular in India. It is mainly used for the social media communication. The authors [10] had created a Hinglish corpus which contained movie reviews domain specific Hindi words. Stop-word removal, tokenization were the pre-processing techniques used in the system, along with TF-IDF as the vectorization technique. Classification algorithms like SVM and Naïve Bayes where used to carry out the classification task. As a future work, the authors in \cite{kaur2017dictionary} are trying to find the best feature and classifier combination.

SVM is the machine learning algorithm which is among the top choice by researchers nowadays. The researchers have even compared the results of the different deep learning models with SVM Sun et al. \cite{sun2018tibetan}. In \cite{sun2018tibetan} SA task performed on Tibetan microblog. Word2vec was the vectorization technique used. It converts the words into the numeric vector. After the vectorization step the classification of the data was carried out by the different machine learning and deep learning algorithms like SVM, Convolution Neural Network (CNN), Long short-term memory (LSTM), CNN-LSTM. CNN is a type of neural network having 4 layers. Input layer, convolution layer, global max pooling layer, output layer. Convolutional layer is the main layer because feature extraction is done in this layer. LSTM is the variant of the RNN (Recurrent Neural Network) which are capable of learning long term dependencies and detecting patterns in the data. The comparative study of different algorithm displays CNN-LSTM model as the best model with the accuracy of 86.21\% \cite{sun2018tibetan}. 

Joshi et al. \cite{joshi2017approach} carried out SA on the Gujarati tweets. Stopword removal, stemming were the pre-processing techniques used in the implemented model. Feature extraction technique Parts of Speech (POS) tagging and the classification algorithm SVM was used in the system. SVM performed very well and gave the accuracy of 92\%. Sharma et al. \cite{sharma2016prediction} tried to predict the Indian election results by extracting the Hindi tweets for political domain. The tweets were mainly for 5 major political parties. Three approaches where implemented to predict the winner in the election. First approach was dictionary based in which n-gram was used as a pre-processing technique and TF-IDF was used as a vectorization technique. SWN was used to classify the data and assign the polarity score to the words. Naïve Bayes algorithm and SVM were the remaining two approaches which were used. SVM and Naïve Bayes predicted party BJP (Bhartiya Janta Party) as the winner. SVM had the accuracy of 78.4\% which was highest among the three implemented approaches.

The authors, Phani et al. \cite{phani2016sentiment} carried out SA in three different languages Hindi, Tamil and Bengali. Feature extraction techniques n-grams and surface features were explored in detail because they were language independent, simple and robust. 12 surface features where considered in the study in which some of them were number of the words in tweet, number of hashtags in the tweet, number of characters in the tweet etc. Comparative study was carried out to find out which feature extraction and sentiment classifier algorithm worked best together. The classifiers like Multinomial Naïve Bayes, Logical Regression (LR), Decision Trees, Random Forest, SVM SVC and SVM Linear SVC were applied on the dataset. Majority of the languages worked best with the word unigram and LR algorithm. Highest accuracy of 81.57\% was for Hindi \cite{phani2016sentiment}. Research by Sahu et al. \cite{sahu2016sentiment} was carried out on movie reviews in Odia language. Naïve Bayes, Logistic Regression, SVM were used for the purpose of classification. Comparison of the results of different algorithms was done using performance metrics like accuracy, precision and recall. Logistic Regression performed the best with the accuracy of 88\% followed by Naïve Bayes with accuracy of 81\% and SVM with the accuracy of 60\% \cite{sahu2016sentiment}.

In paper by, Guthier et al. \cite{guthier2017language} proposed the language independent approach for SA. An emoticon dictionary was created and score were assigned to the emoticons. When the tweet contained the combination of Hashtags and emoticon, The hashtags were also added in the dictionary. A graph-based approach was implemented in the study. The graph-based approach worked on the principle, if multiple hashtags were present in the sentence then all the hashtags would have the same sentiment score. Also, all the hashtags present in the same sentence could be linked with each other. The work was tested on 5 different languages and the accuracy obtained was above 75\%. Average accuracy of the model was 79.8\%. The approach worked fairly with the single word hashtags and the hashtags which formed the sentences and accuracy for them were 98.3\% and 84.5\% respectively.

Kaur et al. \cite{kaur2019cooking} worked on the Hinglish language dataset. YouTube comments of two popular cookery channels were extracted and analysis was carried on them. Pre-processing techniques like stop words removal, null values removal, spell errors removal, tokenization and stemming were performed. DBSCAN which is the unsupervised learning clustering algorithm was used and 7 clusters were formed for the entire dataset. Dataset was manually annotated with the labels of 7 classes. 8 machine learning algorithms were used to perform the sentiment classification. Logistic regression along with term frequency vectorization outperforms the other classification techniques with the accuracy of 74.01\% for one dataset and 75.37\% for the other dataset. Statistical testing was also being carried out to confirm the accuracy of the classifiers.

Both document level and sentence level SA extract the sentiments for the given text but the feature for which the sentiment is expressed cannot be found out. This shortcoming is fulfilled by aspect level SA.

\subsection{Aspect Level}

Aspect level SA is carried out in two steps. First step is to find the features or the components in the text and the second step is to find polarity of sentiments attached to each feature. For e.g. Mobile reviews are given in the series of the tweets. The companies first find out which part or feature of the mobile the users are talking about and then find out the emotions related to that feature.

In the paper by Ekbal et al. \cite{akhtar2016aspect} the aspect level SA was carried out on the product reviews. Dataset was obtained by web scrapping on different websites. Multi-domain product reviews obtained were analyzed in two steps process, first step was aspect extraction i.e. the aspects(features) in the review were extracted using the Condition Random Field Algorithm. In the second step SVM was used to carry out the SA task. Performance evaluation metrics like F-measure and accuracy were used. SVM gave the accuracy of 54.05\% for sentiment classification.

The proposed work by Ray et al. \cite{ray2019mixed} is SA of twitter data. POS tagging was used as feature extraction technique. Word embedding was used as the vectorization technique. Word embedding is the method where the words of sentences are converted into vectors of real numbers. Aspect were not directly labelled instead aspects were tagged to predefined list of categories. Classification of the data was done using three approaches CNN, Rule based approach, CNN + Rule based approach. The hybrid model of CNN + Rule based approach gave the accuracy of 87\%.
Table 1 is the representation of the work done by different researchers in indigenous language. 

\begin{table}[htbp]
\centering
{\begin{tabular}{ | m{4em} | m{1.8cm}| m{3cm} | m{3cm} | m{2.5cm} |}  

\hline 
Dataset & Indigenous Language & Methodology & Results & Authors
\\
\hline

 Twitter & Irish & Machine Translation, Irish Lexicon &	Irish Lexicon has 6\% higher accuracy results than Machine Translation & Haithem et al. \cite{afli2017sentiment} 
 \\
 \hline
 & Gujarati & SVM &	Accuracy - 92\%	& Joshi et al. \cite{joshi2017approach}  \\
\hline
 &Hindi	&Dictionary-based approach, SVM and Naïve Bayes	&Predictions for BJP as winner	&Sharma et al. \cite{sharma2016prediction}\\
\hline 
 & Arabic , English, Spanish, French, German	& Dictionary based approach	& Accuracy - 79.8\%	& Guthier et al. \cite{guthier2017language} \\
 \hline
 & Hindi, Tamil, Bengali&	Multinomial Naïve Bayes, Logical Regression (LR), Decision Trees, Random Forest, SVM SVC, SVM Linear SVC & Accuracy of Hindi 2 class classification - 81.57\%&	Phani et al. \cite{phani2016sentiment} \\
 \hline
 & English 	& CNN and rule-based approach &	Accuracy - 87\%	& Ray et al. \cite{ray2019mixed} \\
 \hline
 & English	& Naïve Bayes, SVM-R, SVM-P, BLM, MLP, LSTM, Bi-LSTM, CNN & Accuracy - 90\%	&Khatua et al. \cite{khatua2019tweeting} \\
\hline
& Spanish	& Machine Learning techniques, FastText Classifier, BERT Classifier & F-measure -45\% & Godino et al. \cite{godino2019gth}\\
\hline
Movie Reviews &	Hindi &	Naive Bayes, POS tagging &	Accuracy - 87\% &	Jha et al. \cite{jha2015homs} \\
\hline
& Hindi	& HNSW, Random Forest, SVM 	& Accuracy - 91\%	&Nanda et al. \cite{nanda2018sentiment}\\
\hline
& Odia	& Naive Bayes, Logical Regression, SVM	& Accuracy - 88\%	&Sahu et al. \cite{sahu2016sentiment}\\
\hline
& Hindi	& CNN	& Accuracy - 95\%	& Rani et al. \cite{rani2019deep}\\
\hline
Web scrapping &Hindi	&Neural Network, HNSW	& Accuracy - 71.5\%	&Yadav et al. \cite{yadav2019semi} \\
\hline
& Hindi and Marathi	& HNSW, English SWN, Random Forest, Naive Bayes, SVM & Marathi Language Accuracy - 90\%, Hindi Language Accuracy- 70\%	&Ansari et al. \cite{ansari2018sentiment} \\
\hline
&Hindi &Condition Random Field, SVM	& Accuracy - 54.05\%	&Ekbal et al. \cite{akhtar2016aspect}\\
\hline
Manually collected speeches	&Hindi&	HNSW + LM Classifier &	Defined domains using the hybrid model	&Pundlik et al. \cite{pundlik2016multiclass} \\
\hline
e-newspapers&	Telugu&	Telugu SWN&	subjectivity classification Accuracy - 74\%, sentiment classification Accuracy - 81\% &	Naidu et al. \cite{naidu2017sentiment}\\
\hline
Tibetan microblog	&Tibetan& 	CNN, SVM, LSTM, CNN-LSTM&	Precision - 86.21\%	&Sun et al. \cite{sun2018tibetan} \\
\hline
YouTube comments	&Hinglish&	DBSCAN, Decision Trees, RF, Naive Bayes, SVM & Accuracy - 74.01\%, Accuracy - 75.37\% &	Kaur et al. \cite{kaur2019cooking} \\
\hline
FIRE 2015	&Hindi, Bengali, Tamil, Telugu&	Logical Regression, SVM, Language specific SWN's&	8\% better accuracy results compared to previous system&	Bhargava et al. \cite{bhargava2016sentiment} \\
\hline
\end{tabular}}
\caption{\label{tab:table-name}Review Papers.}
\end{table}

\section{Current Trending Techniques in NLP}

The traditional machine learning and lexicon-based approaches did not give the expected results. With the emergence of the deep learning techniques like CNN, RNN, LSTM the performance improvements in the results was observed. The main problem of the deep learning algorithms is that they have high complexity and computational cost. BERT, ELMo are few pre-trained classifiers which solved the problems of the deep learning models and also outperformed them. This section identifies the different papers in which deep learning models and advanced models like BERT, ELMo etc. are used. 

In the paper by, Hoang et al. \cite{hoang2019aspect} aspect-based sentiment analysis on the SemEval-2016 - Task 5 was performed. There were three models implemented in the paper, the aspect classification model which identified whether the aspect was related or not to the text. Sentiment Classifier which classified the text into the three sentiment classes positive, negative, neutral. Both of the classifiers follow the structure of the sentence pair classifier which takes two inputs, the classifier token and the separation token which were added to the beginning and end of the sentences respectively. Final classifier implemented was the combined model which identified the sentiments of the text as well as the aspect of the text. The sentence pair classifier is the part of the Bidirectional encoder representation from transformer (BERT) model. BERT is a bidirectional and unsupervised language representation model. It considers the context of a word from both left to right and right to left simultaneously and provide better features compared to the traditional models. The performance of the combined model was better than the traditional approaches and was tested on 18 different datasets.

Khatua et al. \cite{khatua2019tweeting} performed SA on the twitter to understand user’s response on the supreme court verdict of the decimalization of the LGBT. The authors had extracted 0.58 million tweets and used different machine learning and deep learning classifiers like Naïve Bayes, SVM-R, SVM-P, BLM, multi layer perceptron (MLP), Long short-term memory (LSTM), Bi- LSTM and CNN. Bi-LSTM are special type of LSTM in which the information is available from forward to backward and backward to forward that is in both directions. Bi – LSTM outperforms with the accuracy of 90\%.  

In this study, Rani et al. \cite{rani2019deep} have performed SA on the Hindi movie reviews collected from e-newspapers and different online websites. The classification technique used in the paper was CNN. CNN gave the accuracy of 95\% which was much higher than the other traditional algorithms.

In the paper, Godino et al. \cite{godino2019gth} carried out SA on Spanish tweets using three different classifier models which are feature classifier, FastText classifier, BERT classifier. Feature classifier extracted the important features from the tweets such as the length of the tweets, number of hashtags etc. and applied these features to the traditional machine learning algorithms to carry out the sentiment classification. The traditional algorithms used where: Logistic Regression, Multinomial Naive Bayes, Decision Tree, Support Vector Machines, Random Forest, Extra Trees, AdaBoost and Gradient Boost. FastText Classifier was developed by Facebook AI research and it internally works on the neural network architecture. BERT Classifier was also applied on the tweets. The output of the three classifiers were combined using the averaging assembling. The model was evaluated using the F1 score. F1 score of 45\% and 46\% was obtained on the train and test data of the implemented model.

\section{Datasets}

With the increasing use of the web there is a lot of User Generated Content (UGC) available on different websites. Lot of research is carried out for the English language. Work done for the indigenous languages is less as compared to the English language. By studying different papers on SA, it can be found out that researchers have started working on the indigenous languages. Data for the indigenous languages is available across the web but is mainly collected from social media platforms like Twitter, Facebook and YouTube. 

Some researchers have extracted their data from Twitter \cite{afli2017sentiment, joshi2017approach, sharma2016prediction, guthier2017language, ray2019mixed, khatua2019tweeting, godino2019gth}, while some have opted for extracting the data manually or by performing web scrapping on different websites like Facebook, microblogs, e-commerce websites, YouTube etc. \cite{ansari2018sentiment, jha2015homs, nanda2018sentiment, pandey2015framework, kaur2017dictionary, akhtar2016aspect}.  Authors in \cite{bhargava2016sentiment} have accessed the FIRE 2015 dataset. The dataset has 792 utterances and has 8 different languages other than English. Researchers in \cite{sahu2016sentiment} collected 3000 positive and 3000 negative Odia movie reviews. Authors in \cite{naidu2017sentiment} collected 1400 Telugu sentences from e-Newspapers from data 1st December 2016 to 31st December 2016. 

The study in \cite{pundlik2016multiclass} contained the speeches of different leaders who spoke about different domain topics like festivals, environment, society etc. The dataset was manually created. \cite{sun2018tibetan} performed SA on the Tibetan language and hence collected the data from the Tibetan micro-blog. In \cite{yadav2019semi} 112 Hindi text file pertaining to different domains have been collected for analysis. Authors in \cite{phani2016sentiment} have used the SAIL Dataset which consist of training and test data for three different languages. Approximately 1000 tweets for each language was present as a training data. \cite{kaur2019cooking} extracted the data from the YouTube comments. The data extracted was related to the cookery website from 2 channels. Total of 9800 comments were collected.

Major observations made in this paper are listed below. Not many researches have carried out SA on the large dataset, Majority of the research work is done on Facebook, Twitter, YouTube data, Extensive research is mainly carried out only on 2 domains which are movie reviews and politics. Very few researches are done on the cookery websites, medical data, multi-domain data. Data is not extracted from the popular social media platforms like Instagram, LinkedIn in spite of Instagram and LinkedIn being among the top websites used by the people.

\section{Classification Techniques}

Sentiment Analysis is the Natural language processing task. Machine Learning, Deep learning and Lexicon based approach are mainly used to classify the data based on the sentiments. Rule based approaches which were once used for the SA task are now used to carry out the pre-processing and feature extraction on the data. 

Machine learning based approaches split the data into the training and test set. The training set trains the different machine learning algorithms so that they can understand the patterns present in the data and helps in finding the association between the different attributes in the data which can further help for future predictions. After the machine learning algorithms are trained the test set helps the algorithm to check the accuracy of the model. Accuracy helps us to understand how much the algorithm was able to learn from the training set and perform on the unknown data (test set). In the lexicon-based approach the words present in the dataset are searched in the SWN’s. Lexicon based approach is considered as an unsupervised learning technique because it does not require any prior knowledge about the data. Rule Based approaches are approaches which have a set of rules which are to be applied to the dataset to carry out the SA task. 

In various studies machine learning algorithms were used to carry out the SA task \cite{ansari2018sentiment,jha2015homs, nanda2018sentiment,joshi2017approach, sahu2016sentiment, kaur2019cooking,akhtar2016aspect}. It was observed that SVM performed very well for the sentiment classification followed by LR and Naïve Bayes algorithm. Deep learning algorithms like CNN, LSTM, Bi-LSTM were applied on the datasets to find out the performance improvement over the traditional machine learning algorithms. From the final analysis it was concluded that the CNN-LSTM and Bi-LSTM performed the best as compared to the other algorithms \cite{sun2018tibetan, ray2019mixed, khatua2019tweeting,donthula2019man}. 

In some paper’s lexicon-based approach was used to carry out the classification task \cite{afli2017sentiment, naidu2017sentiment, pandey2015framework, kaur2017dictionary, phani2016sentiment, guthier2017language}. SWN’s of different languages were created and improved to carry out the task effectively. Some studies suggested use of both Lexicon and Machine Learning approaches to carry out SA task. Also, suggestions to compare the algorithms and find the best algorithm was given by \cite{pundlik2016multiclass, yadav2019semi,sharma2016prediction}. In \cite{bhargava2016sentiment} Machine learning algorithms LR and SVM were used for Language detection and SWN was used for classification of sentiments. SVM outperformed LR in Language detection. 

With the advancement of techniques various advanced deep learning algorithms like BERT, ELMo, FastText Classifier were applied on the datasets BERT classifier performed the best \cite{hoang2019aspect,godino2019gth}. Different rule-based approach has been used for pre-processing of the data because without the pre-processing the accuracy of the model cannot be found out correctly. 

\section{Challenges and Limitations}

The main challenges faced by the authors are the availability of the annotated corpora, poor quality SWNs or no SWNs, no stop word list for languages. Along with these challenges some of the individual specific challenges faced by the authors are listed below. In \cite{pundlik2016multiclass} Document having more than 1000 and less than 500 words could not be classified by the implemented model. Ontology was also manually created which can affect the accuracy of the system. In \cite{nanda2018sentiment} the data was classified based on only 2 sentiments positive and negative. Neutral polarity was not considered which could affect the analysis to greater extent. In \cite{bhargava2016sentiment} transliteration of words caused issues. Authors in \cite{kaur2017dictionary} faced issue in automatic detection of the topic hashtags because the context was no provided to the system. In \cite{akhtar2016aspect} Multi word aspect terms were not detected and the accuracy of the negative class was low.

\section{Discussions and Analysis}

After the detailed review of different papers, few points that can be considered for discussion further are mentioned below.

\begin{itemize}
    \item Small Dataset: \\
    There is no substantial research carried out for the sentiment analysis in indigenous language for larger dataset. All the datasets have size in between 10k-20k. Usually the data available on the internet is of millions and millions of rows and hence the models which are not tested on the larger dataset can have accuracy problems. 
    \item Less Usage of Deep Learning Algorithms: \\
    Majority of the research carried out for indigenous languages is performed using Machine Learning algorithms except the research carried out by the authors in \cite{pandey2015framework,khatua2019tweeting,rani2019deep,godino2019gth}. Deep learning algorithms have time and again proved to be much better than the traditional machine learning techniques. 
    \item Non-Availability of corpus: \\ 
    The datasets for many of the indigenous languages are not available easily. Many of the researches have to manually collected the data and hence this becomes one of the reasons for the smaller dataset. 

    \item Non-Availability of the SWNs and WordNet’s: \\ 
    There are lot of Indian Languages which don’t have the WordNet’s and SWNs developed hence some of the researchers had to create the WordNet’s and SWN manually. Also, WordNet’s and SWNs are constantly in the evolving state and are not stable.

\item Code-Mix Languages: \\
There is a lot of code-mix language used especially in India on the social media. As multiple languages are mixed it takes large computation time to first perform the language identification and second perform the SA task. There are no resources like WordNet’s, POS Taggers etc. for the code-mix languages. Hence the research in such languages is limited and still evolving.

\item Less Development on the Aspect Level SA: \\
There are very few research papers available on the SA at the aspect level on the indigenous languages.
\end{itemize}

\section{Conclusion and Future Scope}

In this review paper, the main aim is to understand the recent work that has been done in SA for indigenous languages. 23 papers are being studied to find the trends in the field of SA. 67\% of the papers reviewed have used Machine learning, deep learning and advanced deep learning algorithms. Only 29\% of researchers have used lexicon-based approach. SVM (Support Vector Machine) and LR (Logical Regression) performed the best among the machine learning approach. CNN performed the best in the deep learning techniques and BERT was the choice by the researchers in the advanced deep learning techniques. The code-mix languages are the new non official language which we can see on the web. There isn’t much work done on code-mix language data. Also, a lot of work is done in SA of Hindi language as compared to the other Indian languages like Gujarati, Marathi, Telugu. There is a lot of work carried out in the sentence level of sentiment analysis. There is a need for more SA work to be carried out at document level or aspect. Also, there are very few papers which have multi domain dataset. In majority of the papers, analysis is carried out on the movie reviews and the political domain data. There is a need for research on the other domains like festivals, development, education, sociology etc. Also, there is negligible research done on the data collected from Instagram and LinkedIn. BERT model can be considered for classification of code-mix languages because there has been no such research carried out so far.

The future work will involve the investigation on using the advance deep learning model such as Bert in mix code language classification. We have collected over 20000 reviews (combination of Marathi and English). We would be comparing the state of the art methods discussed in the current paper during our investigation and discussed the insightful.


\bibliographystyle{unsrt}  
\bibliography{references}  


\end{document}